\documentclass[twocolumn,switch]{article}
\usepackage{preprint}

\usepackage[utf8]{inputenc}
\usepackage[T1]{fontenc}
\usepackage{graphicx}
\usepackage{amsmath,amssymb}
\usepackage{multirow}
\usepackage{makecell}
\usepackage{booktabs}
\usepackage{microtype}
\usepackage{tcolorbox}
\tcbuselibrary{breakable}

\usepackage[numbers,square]{natbib}
\usepackage[colorlinks=true,linkcolor=purple,urlcolor=blue,citecolor=cyan,anchorcolor=black]{hyperref}
\hypersetup{
  pdftitle={From Reasoning Depth to Reasoning Breadth: Evaluating Multi-Point Associative Reasoning in Large Language Models},
  pdfauthor={Si'an Xie, Jiaxun Liu, Biao Yang, Wei Yuan, Fan Yang, Tingting Gao, Ming Wu}
}

\title{From Reasoning Depth to Reasoning Breadth: Evaluating Multi-Point Associative Reasoning in Large Language Models}
\author{%
\textbf{Si'an Xie$^{1,*}$, Jiaxun Liu$^{2,*}$, Biao Yang$^3$, Wei Yuan$^{3,\dagger}$,}\\
\textbf{Fan Yang$^3$, Tingting Gao$^3$, Ming Wu$^{1,\dagger}$}\\[2pt]
\normalfont $^1$Beijing University of Posts and Telecommunications\\
\normalfont $^2$Peking University\\
\normalfont $^3$Kuaishou Technology\\[-1pt]
\normalfont\small $^*$Equal contribution. $^\dagger$Corresponding authors.
}

\begin{document}

\twocolumn[\begin{@twocolumnfalse}
\maketitle

\begin{abstract}
Large language models (LLMs) have made substantial progress on reasoning tasks that require increasingly long and complex inferential chains. This progress primarily reflects \textit{reasoning depth}. A complementary and comparatively unexamined capability is \textit{reasoning breadth}: exploring multiple semantic directions in parallel and integrating the resulting clues into one coherent answer. We introduce \textbf{MPAR-Bench}, a bilingual (English–Chinese) benchmark that isolates reasoning breadth through multi-point associative reasoning. Inspired by the cooperative game \textit{Just One}, each item asks a model to recover a hidden target from several independently generated, semantically diverse clues. We construct 1,000 items using a multi-agent clue-generation pipeline, embedding-based diversity filtering, and human verification. Crucially, only the answer space is drawn from public word lists, whereas every clue set is generated from scratch. Beyond exact-match accuracy, we evaluate models along a coarse-to-fine protocol (accuracy, ANLS, embedding similarity, and reasoning-trace verification) and a perturbation suite that probes four distinct robustness axes: clue masking, order shuffling, distractor injection, and multi-step clues. Across evaluated models, perturbations reduce accuracy by 9–18 points in English and 5–12 points in Chinese. Enabling thinking mode improves standard-setting accuracy—especially in English—but does not consistently reduce sensitivity to perturbations, and case-level trace analysis shows that extended reasoning can overturn an initially correct hypothesis. These results indicate that greater reasoning depth does not automatically confer robust reasoning breadth, and that breadth is a measurement axis that current benchmarks largely leave uncovered.
\end{abstract}
\vspace{0.35cm}
\end{@twocolumnfalse}]

\section{Introduction}

\begin{figure}[!t]
\centering
\includegraphics[width=\linewidth]{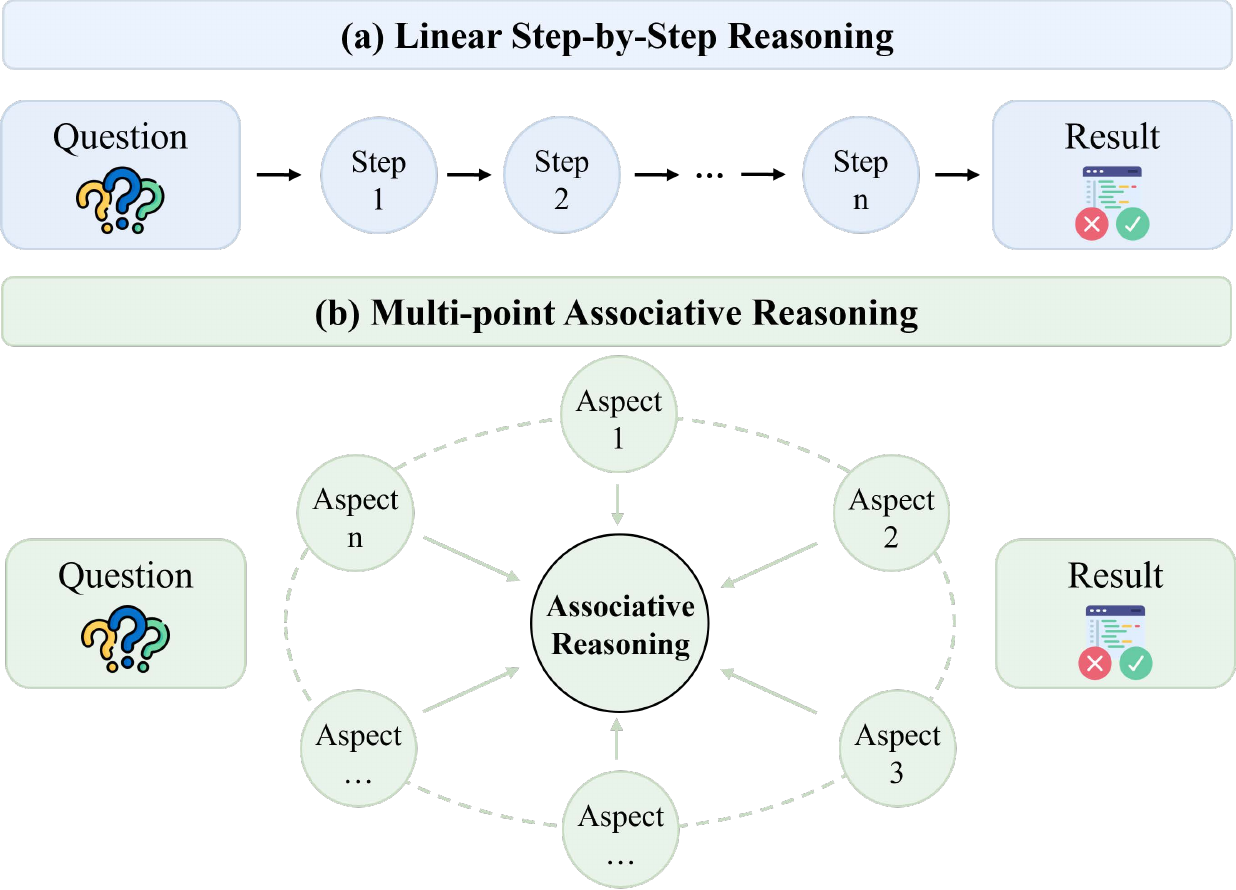}
\caption{Linear Reasoning and Multi-Point Associative Reasoning Flow Chart.}
\label{fig:linear_and_associative}
\end{figure}

Large language models (LLMs) have rapidly evolved from early neural language models to highly capable Transformer-based systems \citep{vaswani2017attention} such as GPT, Gemini, Qwen, and others \citep{openai2026gpt52, google2025gemini3flash, google2026gemini31pro, anthropic2026claude45, yang2025qwen3, liu2025DeepSeek, team2025kimi, bytedance2026seed20}. Through techniques such as reinforcement learning \citep{ouyang2022training}, supervised finetuning \citep{wei2022finetuned}, Chain-of-Thought (CoT) \citep{wei2022chain} and Retrieval-Augmented Generation \citep{lewis2020retrieval}, modern LLMs have achieved remarkable success and demonstrated near-human performance in solving practical problems, which depends on their step-by-step linear reasoning. Humans possess another important capability: multi-point associative reasoning, which enables the structured integration of a wide range of concepts. This ability allows humans not only to explore problems in depth, but also to bridge existing concepts and synthesize them into novel ideas. These ability differences are shown in Fig.~\ref{fig:linear_and_associative}. An important question remains unresolved: do current LLMs possess the non-linear, cross-domain associative reasoning abilities?

\begin{figure*}[!t]
\centering
\includegraphics[width=\textwidth]{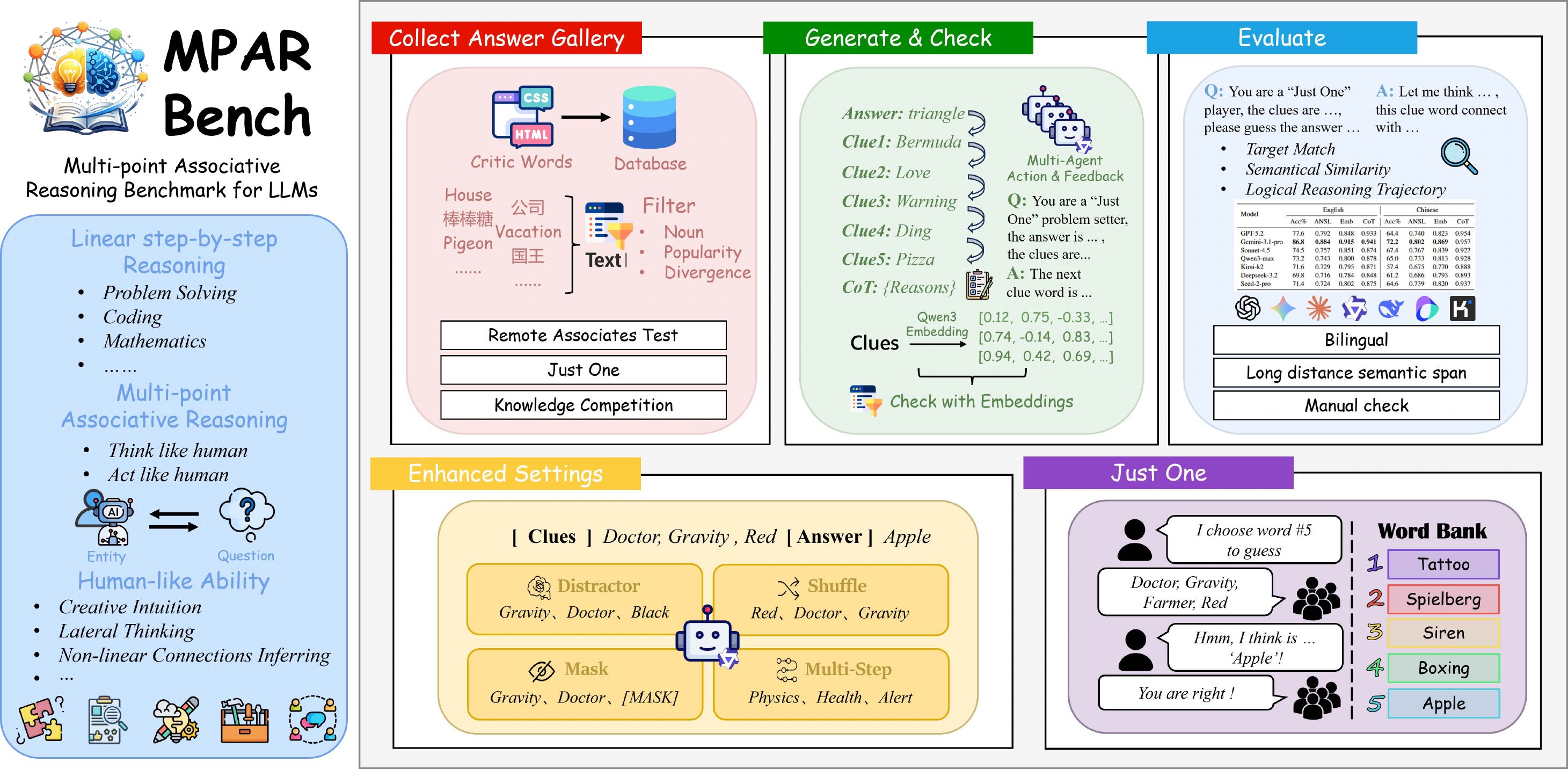}
\caption{A brief introduction of MPAR-Bench.}
\label{fig:total}
\end{figure*}

Existing LLM benchmarks \citep{hendrycks2021math, hendrycks2021measuring, cobbe2021training, srivastava2023beyond, liang2023holistic} primarily emphasize reasoning ``depth,'' evaluating step-by-step logical deduction and procedural reasoning. In contrast, the evaluation of reasoning ``breadth''---the ability to aggregate dispersed semantic signals and perform abstract conceptual convergence---remains largely unexplored. This capability matters whenever the relevant evidence is distributed across different semantic perspectives rather than arranged as a single derivation. Multi-document synthesis, cross-domain analogy, hypothesis generation, and reasoning under incomplete or distracting evidence all require a model to hold several partial relations in view and reconcile them into a final prediction \citep{treutlein2024connecting}. A model may therefore reason deeply along one path while still failing to combine information available across several paths.

To fill this gap, we introduce \textbf{MPAR-Bench}, a cognitively inspired benchmark designed to systematically evaluate multi-point associative reasoning in LLMs. To improve evaluation reliability and robustness, we further construct high-quality bilingual test sets with 1,000 questions through a carefully designed multi-agent generation and verification pipeline \citep{li2023camel, wang2023self, du2024improving}.

Beyond exact-match metrics, we propose a fine-grained evaluation framework that analyzes model behavior from reasoning perspectives. This framework enables a more comprehensive investigation of how LLMs ``reason broad'', offering deeper insight into the current capabilities and limitations of associative reasoning in modern language models. A general overview of MPAR-Bench is shown in Fig.~\ref{fig:total}.

In general, our contributions are summarized as follows:

\begin{itemize}
    \item \textbf{A benchmark targeting reasoning breadth.} MPAR-Bench operationalizes many-to-one integration from multiple, semantically diverse clues. Unlike RAT-style tests (three fixed compound-word cues) and existing game benchmarks (clue giving, or grouping a fixed word set), MPAR-Bench isolates the guesser-side integration of an open number of free-form clues, and pairs it with a controlled perturbation suite.

    \item \textbf{An innovative multi-agent clue-synthesis pipeline}. We propose a multi-agent collaborative clue-synthesis framework with embedding-based filtering and human verification, enabling the construction of semantically diverse, high-difficulty evaluation instances while substantially reducing memorization risk.

    \item \textbf{A coarse-to-fine evaluation protocol.} Beyond exact match, we combine ANLS, embedding similarity, and reasoning-trace verification, and analyze robustness per perturbation type rather than as a single aggregate, exposing a blank space in reasoning breadth.
\end{itemize}

\section{Related Work}

\subsection{LLM in Reasoning Depth}
\label{sec:rw-depth}

The dominant trajectory of LLM reasoning research extends inferential depth. Chain-of-thought prompting \citep{wei2022chain}, tree- and plan-structured search \citep{yao2023tree, wang2023plan}, self-verification \citep{weng2023large}, and process supervision \citep{lightman2024lets} all lengthen or stabilize a single reasoning trajectory, and reinforcement-learned thinking modes push test-time computation further \citep{DeepSeek2025r1}. A parallel line of work documents the failure mode of overthinking, in which additional reasoning steps degrade rather than improve answers \citep{chen2025overthinking, sui2025stop}. Depth-oriented benchmarks---mathematical \citep{hendrycks2021math, cobbe2021training}, knowledge-intensive \citep{hendrycks2021measuring}, and broad-coverage suites \citep{srivastava2023beyond, liang2023holistic}---are increasingly saturated for frontier models. These tasks answer how far a model can push one chain; they do not answer whether a model can integrate evidence across chains. Reasoning breadth is thus an orthogonal axis, and one on which depth-oriented benchmarks provide little discrimination.

\subsection{Associative Reasoning}
\label{sec:rw-assoc}

Associative reasoning has long been studied in cognitive psychology through convergent-thinking instruments, most notably the Remote Associates Test (RAT) \citep{mednick1962associative}. RAT has recently been repurposed to probe LLMs: \citet{schon2022modeling} model associative reasoning processes, \citet{kumar2025human} study human--AI convergent and divergent thinking, and generative models have been reported to match or exceed humans on such tests \citep{carolus2025time, arora2025generative}---though such results are hard to interpret, since the test items are publicly available and may have been seen during pre-training \citep{deng2024unveiling, chen2025benchmarking}.

Related open-ended formulations argue for process-based rather than multiple-choice evaluation \citep{huang2025mm} and for generating explicit associative paths \citep{wadhwa2026create}. We inherit the construct of convergent association from this tradition but deliberately depart from its instrument: RAT items are public, largely three-cue, and dominated by fixed phrasal collocations that next token prediction learns readily. MPAR-Bench instead poses an open, variable number of free-form semantic clues that must be integrated, with every clue set synthesized de novo to substantially reduce overlap with public clue--target pairings and lower the risk of memorization. This is what MPAR-Bench adds over prior RAT-on-LLM evaluations: a breadth-oriented benchmark with reduced contamination risk, rather than a re-run of a public convergent-thinking test.

\begin{figure}[!t]
\centering
\includegraphics[width=\linewidth]{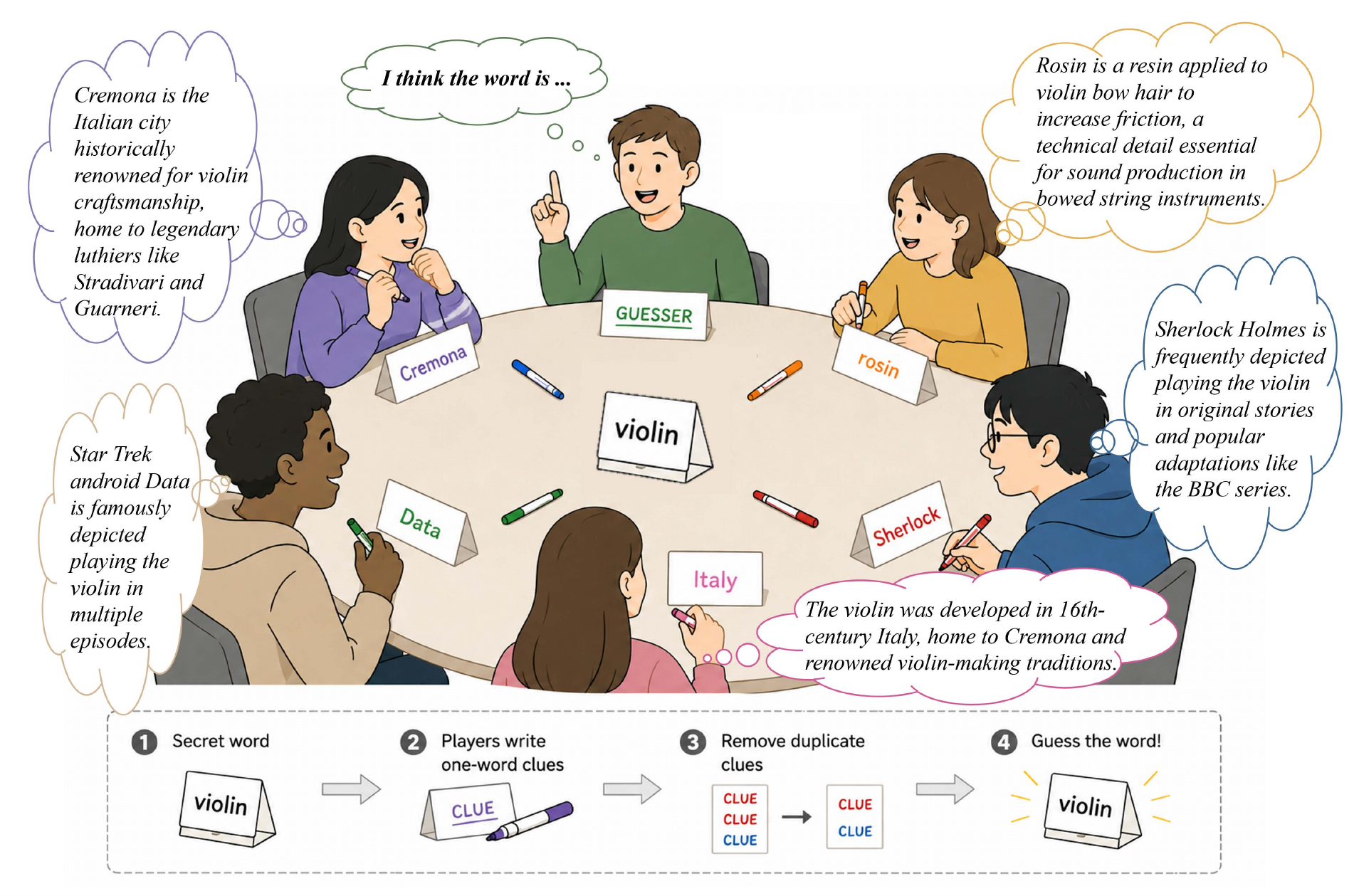}
\caption{Introduction of Board Game \textit{Just One}.}
\label{fig:just_one}
\end{figure}

\subsection{Boardgame-Based Benchmarks}
\label{sec:rw-games}

Cooperative and word-association games offer constrained rules with large state spaces, which helps mitigate contamination and yields human-aligned semantic tasks. Codenames has been used to evaluate one-to-many clue giving and ad-hoc concept forming \citep{stephenson2025codenames, hakimov2025ad}; the NYT Connections game requires partitioning a fixed set of words into latent groups \citep{merino2024making}; and the Word Synchronization Challenge measures two agents converging on a shared word without communication \citep{cazalets2025word}. MPAR-Bench differs along three axes: \textbf{(i) clue cardinality}---models must jointly integrate an open, variable number of clues rather than fixed cues; \textbf{(ii) association type}---clues are free-form semantic descriptions spanning lexical, cultural, phonetic, and world-knowledge relations, not compound-word completions or fixed candidate pools; and \textbf{(iii) item availability}---all clue sets are synthesized from scratch, leaving no public clue--target pairing to memorize.

\section{Methodology}

\subsection{Task Definition}
\label{sec:formulation}

Given a clue set $C = \{c_1, c_2, \ldots, c_n\}$, the task is to recover a target $y$ such that each clue $c_i$ contributes an \emph{independently informative} semantic relation to $y$. Reasoning breadth, in this setting, is the ability to integrate multiple semantically distinct and non-redundant clues into a single coherent answer.

We ensure that each item genuinely requires breadth through two construction-side safeguards. First, clues are generated from diverse semantic angles to maximize the range of associations a model must reconcile. Second, a judge agent and an embedding-based filter remove synonyms, paraphrases, and near-duplicates, so that each retained clue carries non-overlapping information. At evaluation time, we measure not only whether a model's prediction matches the target, but also whether that prediction remains stable under perturbation---clue masking, order shuffling, distractor injection, and multi-step inference---which probes whether the integration is robust or merely superficial.

\begin{figure}[!t]
\centering
\includegraphics[width=\linewidth]{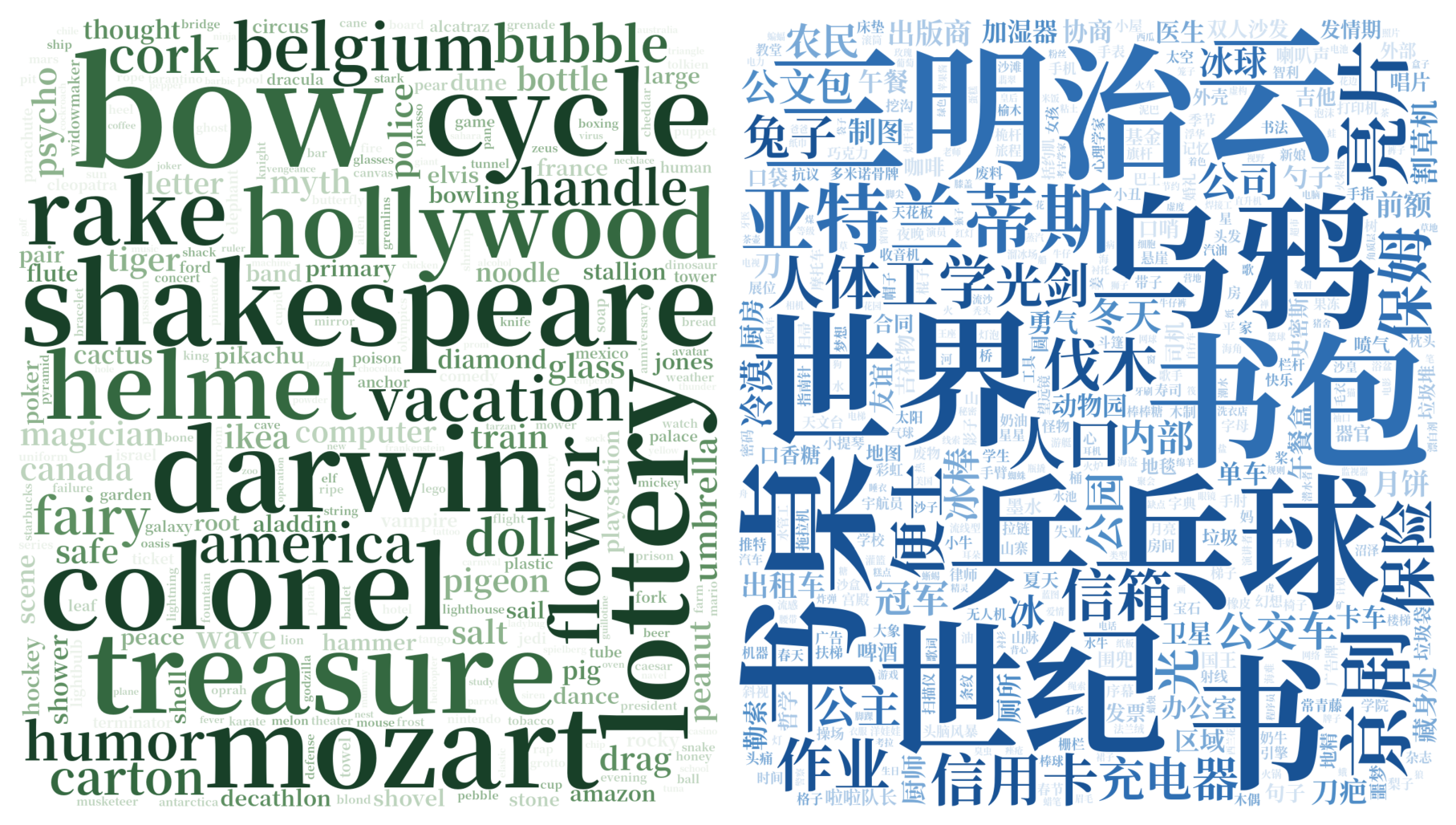}
\caption{Word Cloud of MPAR-Bench.}
\label{fig:word_cloud}
\end{figure}

\subsection{Just One}
\label{sec:justone}

The design borrows the constraint structure of the cooperative game \textit{Just One}, in which players give single-word hints to help a guesser infer a hidden target, while direct synonyms, translations, homophones, and duplicate clues are forbidden. These constraints are what make the game a clean instrument for breadth. Rather than rewarding the most obvious lexical association, they force clue writers to approach the target from distinct, indirect angles; the guesser must then integrate fragmented, non-overlapping signals rather than pattern-match a single cue. This yields a constrained multi-point associative reasoning task emphasizing semantic abstraction, conceptual bridging, and integration. An illustrative round is shown in Fig.~\ref{fig:just_one}.

\subsection{Dataset Construction}
\label{sec:dataset}

\noindent\textbf{Answer space.} Target words are drawn from public word lists---RAT-derived vocabulary (collected on the internet) and \textit{Just One} word cards.

\noindent\textbf{Multi-agent clue generation.} Given a target and the clues already accepted, LLM-based agents iteratively propose new clues that remain semantically relevant to the target while minimizing redundancy with existing clues. Each agent is assigned a distinct association angle to encourage coverage across semantic directions. A judge agent then removes clues that are the answer itself, direct synonyms/translations/homophones/morphological variants, exact or near-duplicates of accepted clues, or genuinely low quality \citep{long2024llms, bavaresco2025llms}. This division of labor mirrors the independent clue-provider and arbiter structure of \textit{Just One}. Notably, all agents' prompt are provided in Appendix.

\noindent\textbf{Embedding-based diversity filtering.} We use Qwen3-Embedding-8B \citep{zhang2025qwen3} to score clue--answer and clue--clue similarity, discarding clues that are either trivially close to the answer or too weakly related to be informative, and clue pairs that are near-duplicates \citep{abbas2021semdedup}. An experimental embedding similarity threshold ranges from 0.3 to 0.8, which serves as a primary filtering step in benchmark construction.

\noindent\textbf{Answer uniqueness and graded acceptability.} A central concern is that a clue set may admit more than one reasonable target. We address this in two ways. (i) \emph{Construction:} the judge stage filters out clue sets that jointly under-determine the target, and problematic items are reconstructed. (ii) \emph{Human verification:} on a randomly sampled subset of 250 items, two master's students majoring in NLP independently assess whether each item admits a unique, unambiguous target. Table~\ref{tab:ans_uniqueness} reports the results with 95\% Wilson CI, indicating that 92.8\% of the items were judged as having a unique answer.

\begin{table}[t]
\centering
\small
\setlength{\tabcolsep}{1.5pt}
\begin{tabular}{cccc}
\toprule
Category & Count & Rate & 95\% Wilson CI \\
\midrule
Unique & 232 & 92.8\% & [88.6\%, 95.4\%] \\
Ambiguity & 18 & 7.2\% & [4.6\%, 11.1\%] \\
\bottomrule
\end{tabular}
\caption{Answer uniqueness with 95\% Wilson CI}
\label{tab:ans_uniqueness}
\end{table}

\noindent\textbf{Bilingual design.} MPAR-Bench includes English and Chinese subsets built with the same pipeline (500 validated items each), combining synthesized and native-speaker-authored samples. The English subset emphasizes lexical and abstract associations; the Chinese subset additionally incorporates idioms, character-level and pictographic properties, and contemporary cultural memes. Word clouds for each subset are shown in Fig.~\ref{fig:word_cloud}.

\noindent\textbf{Benchmark rationale.} MPAR-Bench is intentionally constructed to emphasize long-range, multi-point association rather than lexical overlap or frequent collocations. By synthesizing clues from complementary perspectives while enforcing low redundancy, each item requires integrating sparse and semantically distant evidence into a single target, making successful prediction less dependent on next token co-occurrence patterns and more on associative reasoning.

\begin{figure}[!t]
\centering
\includegraphics[width=0.95\linewidth]{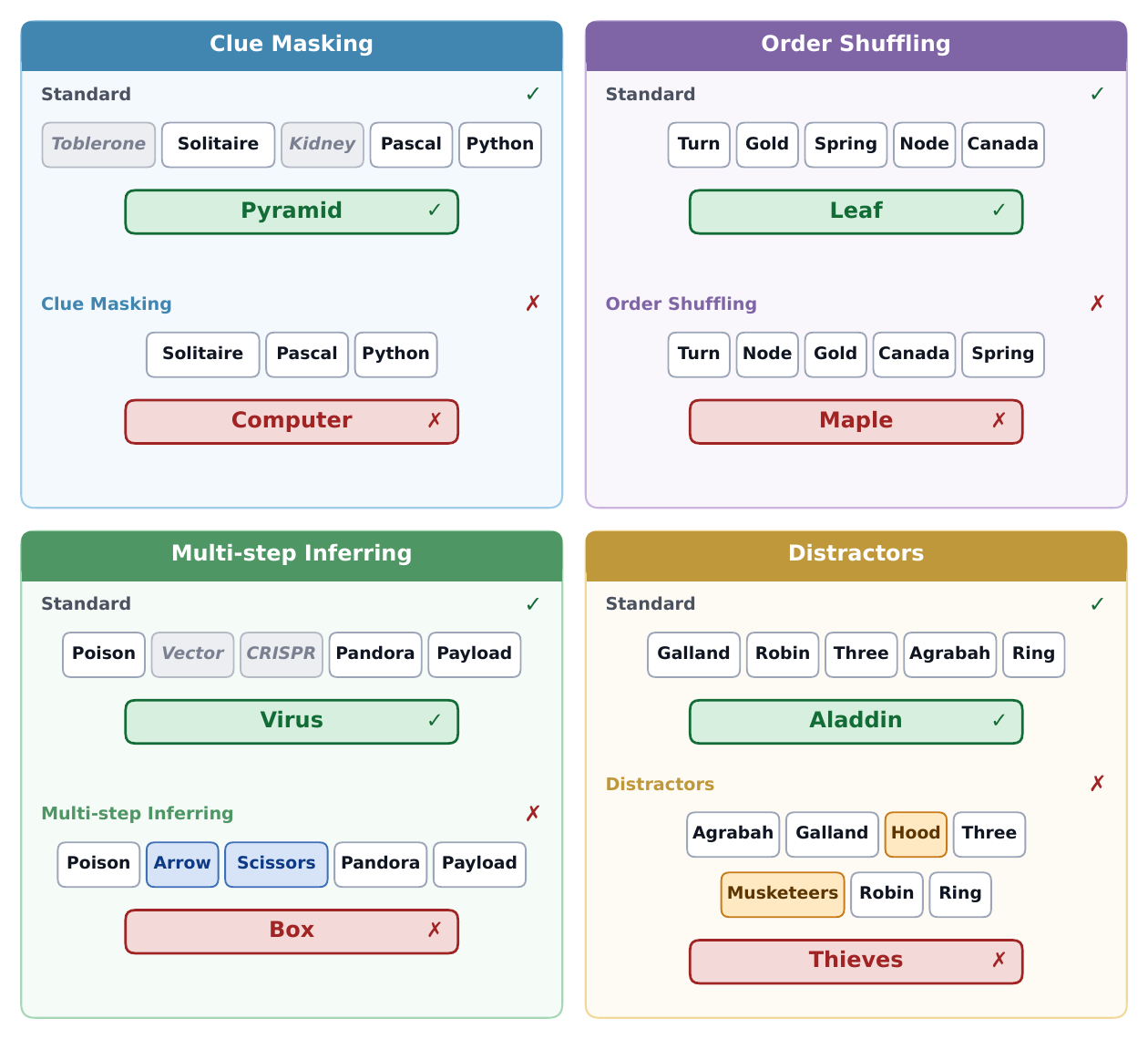}
\caption{Examples of Enhanced settings in MPAR-Bench.}
\label{fig:enhanced_example}
\end{figure}

\subsection{Difficulty Settings and Perturbation}
\label{sec:enhanced}
We distinguish two complementary facets of reasoning breadth. The Standard setting measures baseline breadth: given complete, well-formed clues under ideal conditions, can the model integrate multiple semantic signals into a correct answer? To further evaluate whether this integration capability remains reliable under more realistic conditions, we introduce an Enhanced setting that systematically perturbs the standard test protocol to simulate information-restricted or noisy environments. These two settings provide a complete picture: the Standard setting establishes what a model can achieve under favorable conditions, while the Enhanced setting reveals whether that capability is resilient enough to matter in practice \citep{zhao2025stress}.

Specifically, we implement the following enhanced transformations and perturbations:

\begin{itemize}
    \item \textbf{Clue Masking:} Randomly masking clues to evaluate model reasoning ability under information deficiency.
    \item \textbf{Order Shuffling:} Shuffling the clues order to find whether the model's reasoning process is sensitive to order.
    \item \textbf{Distractors:} Injecting semantically misleading or irrelevant cue words to test model resistance to noisy contexts and spurious correlations.
    \item \textbf{Multi-step Inferring:} Increasing the associative semantic distance between clues and the mystery word, forcing models to generate intermediate latent connections rather than relying on direct surface co-occurrence.
\end{itemize}

Each enhanced setting evenly distributes words from the standard task. We refer to this setting as the Enhanced MPAR-Bench in the remainder of the paper. As shown in Fig.~\ref{fig:enhanced_example}, each question has a corresponding enhanced variant, posing a greater challenge to LLMs across multiple dimensions of robustness.

\section{Evaluation}

Associative reasoning cannot be fully captured by a single exact-match metric, since semantically reasonable predictions may differ lexically from the ground truth, and some correct predictions may come from flawed reasoning processes. To obtain a more comprehensive understanding of LLM behavior, we evaluate models at three progressively finer granularities: accuracy, word-based similarity, and the validity of reasoning trace, i.e., the explanation of how the answer and clues are connecting with each other.

\subsection{Accuracy}
\label{sec:accuracy}

We first evaluate model performance using exact-match accuracy. A prediction is considered correct if and only if it exactly matches the ground-truth answer. We report accuracy across different subsets, including multilingual Standard and Enhanced settings, to measure models' associative retrieval capability under varying semantic and linguistic conditions.

\begin{table}[t]
\centering

\small
\setlength{\tabcolsep}{1pt}

\begin{tabular}{lcccc|cccc}
\toprule
\multirow{2}{*}{Model} &
\multicolumn{4}{c|}{English} &
\multicolumn{4}{c}{Chinese} \\
\cmidrule(r){2-5}
\cmidrule(l){6-9}
& Acc & ANLS & Emb & Trace
& Acc & ANLS & Emb & Trace \\
\midrule
GPT-5.2          & 77.6 & 0.792 & 0.848 & 0.933 & 64.4 & 0.740 & 0.823 & 0.954 \\
Gemini-3.1pro   & \textbf{86.8} & \textbf{0.884} & \textbf{0.915} & \textbf{0.941}
                 & \textbf{72.2} & \textbf{0.802} & \textbf{0.869} & \textbf{0.957} \\
Sonnet-4.5       & 79.0 &  0.803  &  0.856  &  0.865 & 67.4 & 0.767 & 0.839 & 0.927 \\
Qwen3-max        & 73.2 & 0.743 & 0.800 & 0.878 & 65.0 & 0.733 & 0.813 & 0.928 \\
Kimi-k2          & 71.6 & 0.729 & 0.795 & 0.871 & 57.4 & 0.675 & 0.770 & 0.888 \\
Deepseek-v3.2    & 69.8 & 0.716 & 0.784 & 0.848 & 61.2 & 0.686 & 0.793 & 0.893 \\
Seed-2-pro       & 71.4 & 0.724 & 0.802 & 0.875 & 64.6 & 0.739 & 0.820 & 0.937 \\
\bottomrule
\end{tabular}
\caption{Standard MPAR-Bench Results on Thinking Models}
\label{tab:standard_think}
\end{table}
\begin{table}[t]
\centering
\small
\setlength{\tabcolsep}{1pt}
\begin{tabular}{lcccc|cccc}
\toprule
\multirow{2}{*}{Model} &
\multicolumn{4}{c|}{English} &
\multicolumn{4}{c}{Chinese} \\
\cmidrule(r){2-5} \cmidrule(l){6-9}
& Acc & ANLS & Emb & Trace & Acc & ANLS & Emb & Trace \\
\midrule
GPT-5.2 & 66.6 & 0.681 & 0.764 & 0.902 & 58.8 & 0.669 & 0.772 & \textbf{0.932} \\
Gemini-3.1pro & \textbf{76.9} & \textbf{0.778} & \textbf{0.842} & \textbf{0.906} & \textbf{64.0} & \textbf{0.722} & \textbf{0.815} & 0.926 \\
Sonnet-4.5 & 63.4 & 0.646 & 0.739 & 0.836 & 61.0 & 0.689 & 0.789 & 0.873 \\
Qwen3-max & 56.8 & 0.591 & 0.679 & 0.841 & 54.8 & 0.621 & 0.732 & 0.911 \\
Kimi-k2 & 57.6 & 0.596 & 0.687 & 0.827 & 45.8 & 0.547 & 0.683 & 0.850 \\
Deepseek-v3.2 & 52.0 & 0.531 & 0.639 & 0.794 & 52.4 & 0.596 & 0.721 & 0.842 \\
Seed-2-pro & 56.6 & 0.579 & 0.682 & 0.817 & 58.2 & 0.666 & 0.768 & 0.887 \\
\bottomrule
\end{tabular}
\caption{Enhanced MPAR-Bench Results on Thinking Models}
\label{tab:enhanced_think}
\end{table}

\subsection{Word-based Evaluation}
\label{sec:anls}

To further evaluate semantic proximity between predictions and target concepts, we adopt both Average Normalized Levenshtein Similarity (ANLS) and word embedding similarity to judge at the semantic level.

ANLS \citep{biten2019scene} is computed using normalized Levenshtein edit distance:
\begin{equation}
    \text{ANLS}(\hat{y}, y) = 1-\frac{d_{\text{lev}}(\hat{y}, y)}{\max(|\hat{y}|, |y|)},
\end{equation}
where $\hat{y}$ and $y$ denote the model prediction and ground truth respectively, $d_{\text{lev}}(\cdot, \cdot)$ denotes the Levenshtein edit distance, and $|\cdot|$ denotes string length.

Moreover, we compute word embedding similarity using \textit{fastText} \citep{bojanowski2017enriching} as a word embedding model:
\begin{equation}
    \text{Sim}(\hat{y}_{emb}, y_{emb}) = \frac{\hat{y}^\top_{emb} y_{emb}}{\|\hat{y}_{emb}\|_2 \|y_{emb}\|_2}.
\end{equation}
where $\hat{y}_{emb}$ and $y_{emb}$ denote the word embedding of the model prediction and ground truth in \textit{fastText} respectively, and $\|\cdot\|_2$ denotes their Euclidean ($\ell_2$) norms.

\subsection{Reasoning Trace Evaluation}
\label{sec:cot_eval}

Beyond final-answer accuracy, we assess the validity of intermediate processes via reasoning trace evaluation, decomposed into two dimensions: logical verification and factual verification \citep{lanham2023measuring, weng2023large, lightman2024lets}. Logical verification examines whether reasoning trajectories follow coherent inferential steps from clues to predictions, while factual verification checks whether intermediate claims are factually grounded. This dual assessment distinguishes valid associative reasoning from spurious correlations and hallucinated paths. We manually review a randomly sampled subset of 300 reasoning trace predictions, confirming high consistency (98.7\% on factual verification, 94.7\% on logical verification) between human and LLM judgement. Evaluation prompts are provided in Appendix.

\section{Experiments and Results}
\label{sec:experiments_and_results}
\begin{table}[t]
\centering
\small
\setlength{\tabcolsep}{1pt}
\begin{tabular}{lcccc|cccc}
\toprule
\multirow{2}{*}{Model} &
\multicolumn{4}{c|}{English} &
\multicolumn{4}{c}{Chinese} \\
\cmidrule(r){2-5} \cmidrule(l){6-9}
& Acc & ANLS & Emb & Trace & Acc & ANLS & Emb & Trace \\
\midrule
GPT-5.2 & 59.6 & 0.614 & 0.705 & 0.811 & 61.8 & 0.718 & 0.804 & \textbf{0.943} \\
Gemini-3flash & 70.0 & \textbf{0.717} & 0.787 & 0.831 & 67.0 & 0.765 & 0.836 & 0.920 \\
Sonnet-4.5 & \textbf{70.4} & 0.716 & \textbf{0.791} & \textbf{0.838} & \textbf{68.8} & \textbf{0.776} & \textbf{0.843} & 0.921 \\
Qwen3-max & 55.4 & 0.572 & 0.669 & 0.792 & 64.4 & 0.748 & 0.820 & 0.930 \\
DeepSeek-v3.2 & 51.4 & 0.536 & 0.644 & 0.730 & 60.6 & 0.702 & 0.788 & 0.906 \\
Seed-2-pro & 59.8 & 0.619 & 0.711 & 0.743 & 64.2 & 0.726 & 0.817 & 0.910 \\
\bottomrule
\end{tabular}
\caption{Standard MPAR-Bench Results on Non-thinking Models}
\label{tab:standard_non-thinking}
\end{table}

\begin{table}[t]
\centering
\small
\setlength{\tabcolsep}{1pt}
\begin{tabular}{lcccc|cccc}
\toprule
\multirow{2}{*}{Model} &
\multicolumn{4}{c|}{English} &
\multicolumn{4}{c}{Chinese} \\
\cmidrule(r){2-5} \cmidrule(l){6-9}
& Acc & ANLS & Emb & Trace & Acc & ANLS & Emb & Trace \\
\midrule
GPT-5.2 & 44.2 & 0.460 & 0.586 & 0.754 & 54.4 & 0.632 & 0.749 & \textbf{0.906} \\
Gemini-3flash & \textbf{56.0} & \textbf{0.580} & 0.675 & 0.763 & 61.4 & 0.695 & 0.791 & 0.870 \\
Sonnet-4.5 & 55.8 & 0.572 & \textbf{0.679} & \textbf{0.774} & \textbf{62.0} & \textbf{0.707} & \textbf{0.797} & 0.865 \\
Qwen3-max & 44.8 & 0.464 & 0.586 & 0.751 & 56.0 & 0.662 & 0.758 & 0.874 \\
Deepseek-v3.2 & 42.6 & 0.444 & 0.566 & 0.685 & 54.0 & 0.618 & 0.738 & 0.851 \\
Seed-2-pro & 43.4 & 0.452 & 0.586 & 0.693 & 56.0 & 0.646 & 0.761 & 0.866 \\
\bottomrule
\end{tabular}
\caption{Enhanced MPAR-Bench Results on Non-thinking Models}
\label{tab:enhanced_non-thinking}
\end{table}

\subsection{Implementation Details}
We benchmark a diverse set of representative LLM families---including GPT, Gemini, Sonnet, Qwen, Kimi, DeepSeek, and Seed series \citep{openai2026gpt52, google2025gemini3flash, google2026gemini31pro, anthropic2026claude45, yang2025qwen3, liu2025DeepSeek, team2025kimi, bytedance2026seed20}; the concrete information is shown in Appendix. Evaluations are conducted across both thinking and non-thinking modes, under standard and enhanced settings, on the bilingual subsets of MPAR-Bench.

We also report fine-grained metrics including ANLS, word embedding similarity, and reasoning trace failure analysis to dissect the reasoning behaviors of LLMs from lexical, semantic, and reasoning process perspectives.

\subsection{Main Results}

\begin{figure}[!t]
\centering
\includegraphics[width=\linewidth]{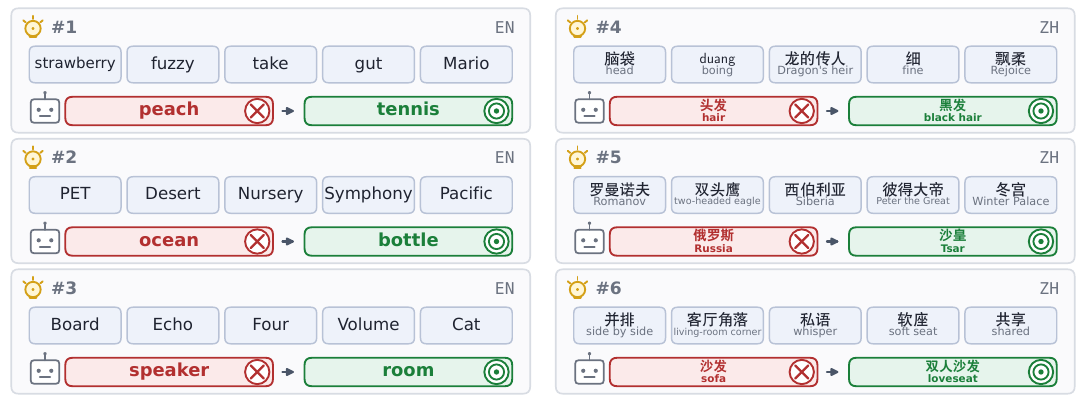}
\caption{Challenging Cases in MPAR-Bench. Examples where most models fail to identify the correct answer.}
\label{fig:result_visualization}
\end{figure}

We analyze the results from three key perspectives: cross-model comparison, the effect of thinking mode, and robustness under perturbation. Challenging cases in MPAR-Bench are shown in Fig.~\ref{fig:result_visualization}, and more detailed experimental results are provided in Appendix.

\paragraph{Model Comparisons.} Table~\ref{tab:standard_think} and Table~\ref{tab:standard_non-thinking} report the standard MPAR-Bench results under thinking and non-thinking modes. Under the thinking mode, Gemini-3.1pro leads on both English (86.8\%) and Chinese (72.2\%), followed by GPT-5.2 and Sonnet-4.5 in English. Under the non-thinking mode, Sonnet-4.5 achieves the highest accuracy in both languages. 

\paragraph{Standard vs. Enhanced.} Comparing standard (Tables~\ref{tab:standard_think}, \ref{tab:standard_non-thinking}) with enhanced settings (Tables~\ref{tab:enhanced_think}, \ref{tab:enhanced_non-thinking}) reveals consistent degradation under perturbation. For instance, in thinking mode, Deepseek-v3.2 exhibits the largest decline in English, while Kimi-k2 drops the most in Chinese. In non-thinking mode, Seed-2-pro shows the largest decline in English, whereas Qwen3-max drops the sharpest in Chinese. These contrasts indicate that robustness varies substantially across models: some suffer pronounced degradation under perturbation while others remain comparatively stable. Notably, this robustness is strongly model- and language-dependent. Detailed information and analyses are shown in Appendix.

\paragraph{Thinking vs. Non-Thinking.} Comparing thinking-mode results (Tables~\ref{tab:standard_think}, \ref{tab:enhanced_think}) against their non-thinking counterparts (Tables~\ref{tab:standard_non-thinking}, \ref{tab:enhanced_non-thinking}), we find that thinking mode \citep{lightman2024lets, DeepSeek2025r1} consistently improves most indicators, but the magnitude of the gain is markedly larger on English than on Chinese: averaged across models, thinking lifts English accuracy by a substantially wider margin and produces clear, stable gains for every model, whereas its effect on Chinese is much smaller and model-dependent (Sonnet-4.5 even shows a slight regression), and the improvement under perturbation is non-monotonic. Moreover, thinking mode is not always reliable: models occasionally overthink and override correct intermediate answers, as discussed in the next section and Appendix. This suggests a crucial distinction: while thinking mode improves reasoning depth, it may not necessarily enhance reasoning breadth.

\section{Discussion}
The benchmark results above establish that current LLMs exhibit measurable but imperfect reasoning breadth, and that this capability degrades under perturbation. A natural follow-up question is: under what conditions does reasoning breadth improve, and what mechanisms cause it to fail? We investigate this question through four complementary lenses. First, we examine overthinking, a failure mode in which extended reasoning actively harms breadth by overriding correct intermediate answers. Second, we characterize information gain curve, asking how breadth scales as more clues become available. Third, we explore scaling laws to determine whether larger models inherently develop broader reasoning. Fourth, we test whether semantic feedback can steer models toward correct answers across multiple refinement rounds. We additionally try to create a structured reasoning skill as an intervention strategy; results are reported in Appendix.

\subsection{Overthinking}
A notable failure pattern we observe is overthinking \citep{chen2025overthinking, sui2025stop}: models initially arrive at the correct answer but subsequently override it during extended reasoning, often drifting toward a semantically related but incorrect concept. This behavior is particularly pronounced in Qwen3-max and Kimi-k2. For instance, given the answer word \textit{Philosophy}, the model outputs \textit{Plato}, over-focusing on a representative entity implied by the clues rather than the academic discipline itself. 

Table~\ref{tab:overthinking_res} reports a detailed overthinking analysis. ``Wrong Ex.'' is the total count of incorrect predictions. ``Ans. Mention'' (ratio) is the proportion of those incorrect cases in which the correct answer appeared in the reasoning trace but was subsequently overridden. ``Token Len.'' (ratio) is the proportion of incorrect cases whose reasoning length exceeded the model's average reasoning length on correct samples.

\begin{table}[t]
\centering
\small
\setlength{\tabcolsep}{1.5pt}
\begin{tabular}{lccc|ccc}
\toprule
\multirow{2}{*}{Model} &
\multicolumn{3}{c|}{English} &
\multicolumn{3}{c}{Chinese} \\
\cmidrule(r){2-4} \cmidrule(l){5-7}
& \makecell{Wrong\\Ex.} & \makecell{Ans.\\ Mention} & \makecell{Token \\ Len.} & \makecell{Wrong\\Ex.} & \makecell{Ans.\\ Mention} & \makecell{Token \\ Len.} \\
\midrule
Sonnet-4.5 & 105 & 0.457 & 0.371 & 163 & 0.706 & 0.571 \\
Qwen3-max & 134 & 0.590 & 0.515 & 174 & 0.874 & 0.690 \\
Kimi-k2 & 142 & 0.585 & 0.415 & 213 & 0.864 & 0.366 \\
Deepseek-v3.2 & 151 & 0.623 & 0.510 & 194 & 0.830 & 0.634 \\
Seed-2-pro & 143 & 0.497 & 0.406 & 177 & 0.751 & 0.599 \\
\bottomrule
\end{tabular}
\caption{Overthinking Results on models. }
\label{tab:overthinking_res}
\end{table}

\begin{figure}[!t]
\centering
\includegraphics[width=0.9\linewidth]{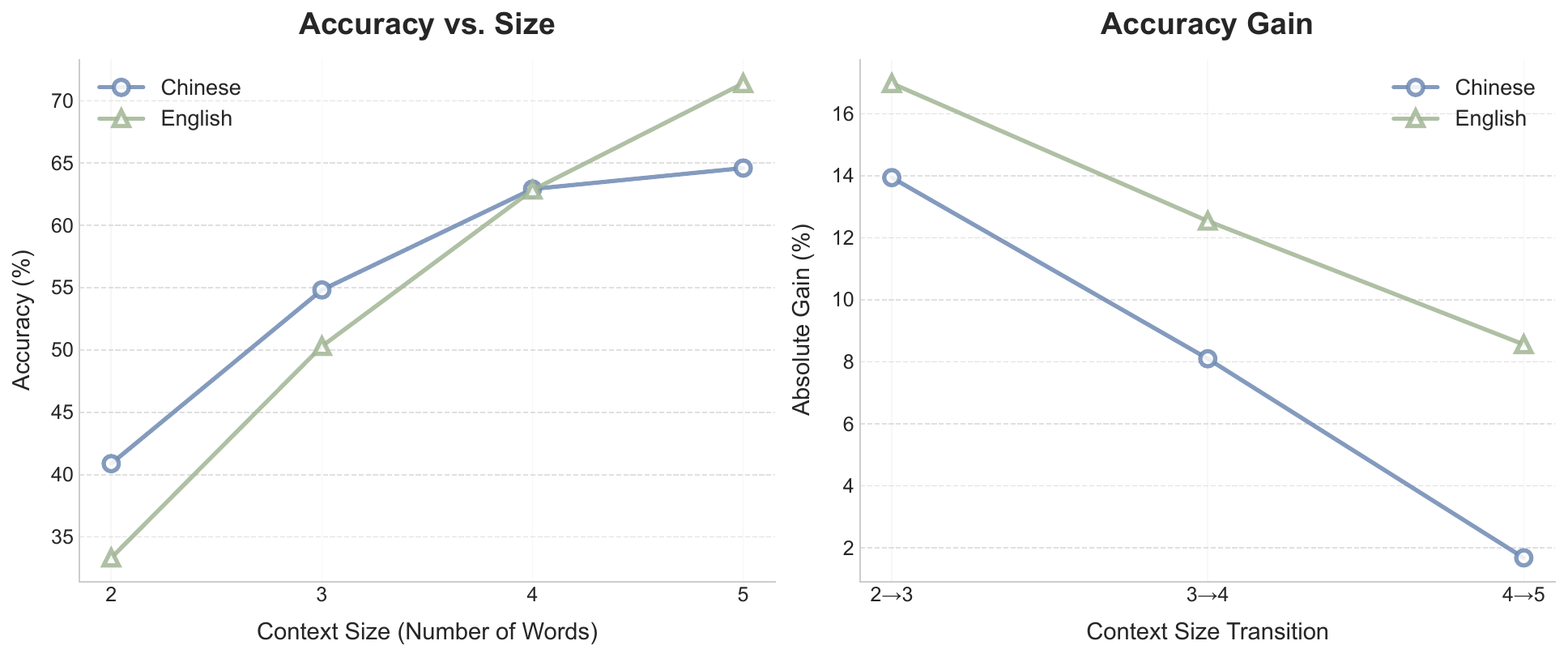}
\caption{Information Gain Curve of Seed-2-pro. As number of words increases, accuracy rises but at a decreasing rate.}
\label{fig:gain_visualization}
\end{figure}

\subsection{Information Gain Curve}
The information gain curve characterizes how a model's reasoning accuracy scales with the number of provided clue words; we utilize it to evaluate Seed-2-pro's capacity to leverage incremental semantic evidence for multi-point associative reasoning. 
As shown in Fig.~\ref{fig:gain_visualization}, accuracy consistently improves as the number of clue words increases, suggesting that the model is able to accumulate and integrate incremental semantic information across multiple clues. This provides preliminary evidence that LLMs possess multi-point associative reasoning capability—the ability to jointly combine several semantically distinct clues into a coherent answer. However, the marginal gain progressively slows down, indicating that while models benefit from richer semantic context, they saturate beyond a certain evidence threshold.

\subsection{Scaling Law}
\begin{figure}[!t]
\centering
\includegraphics[width=\linewidth]{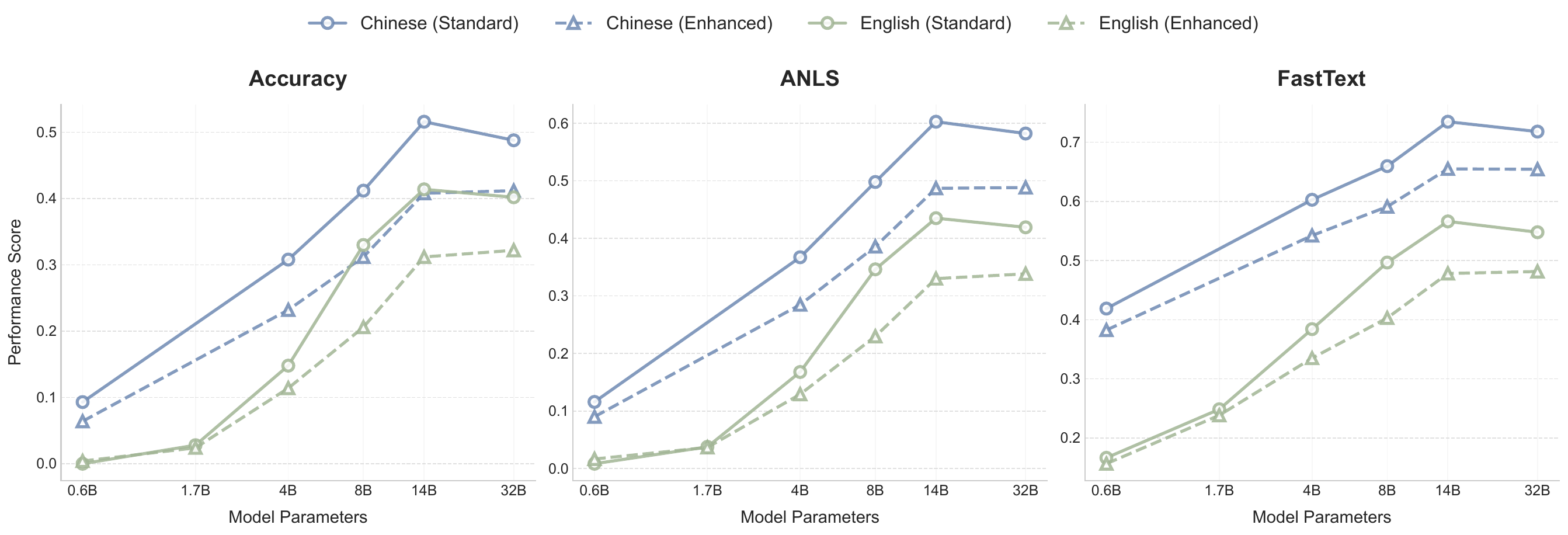}
\caption{Scaling Law of Qwen3 family in Chinese MPAR-Bench. As model scales up, all metrics improve, except for Qwen3-32B, which has a more severe overthinking issue.}
\label{fig:scaling_law}
\end{figure}
We investigate how model scale affects performance on our benchmark. In particular, Fig.~\ref{fig:scaling_law} reports the scaling behavior of the locally deployed Qwen3 family in thinking mode (0.6B to 32B parameters), evaluated on accuracy, ANLS, and fastText embedding similarity across both English and Chinese datasets under Standard and Enhanced settings. Except for Qwen3-32B, accuracy consistently improves with increasing model size \citep{kaplan2020scaling, hoffmann2022training, schaeffer2023emergent}. Case-level inspection finds that Qwen3-32B suffers from overthinking, causing it to reject correct answers during an extended reasoning process.

\subsection{Feedback}
\begin{figure}[!t]
\centering
\includegraphics[width=0.94\linewidth]{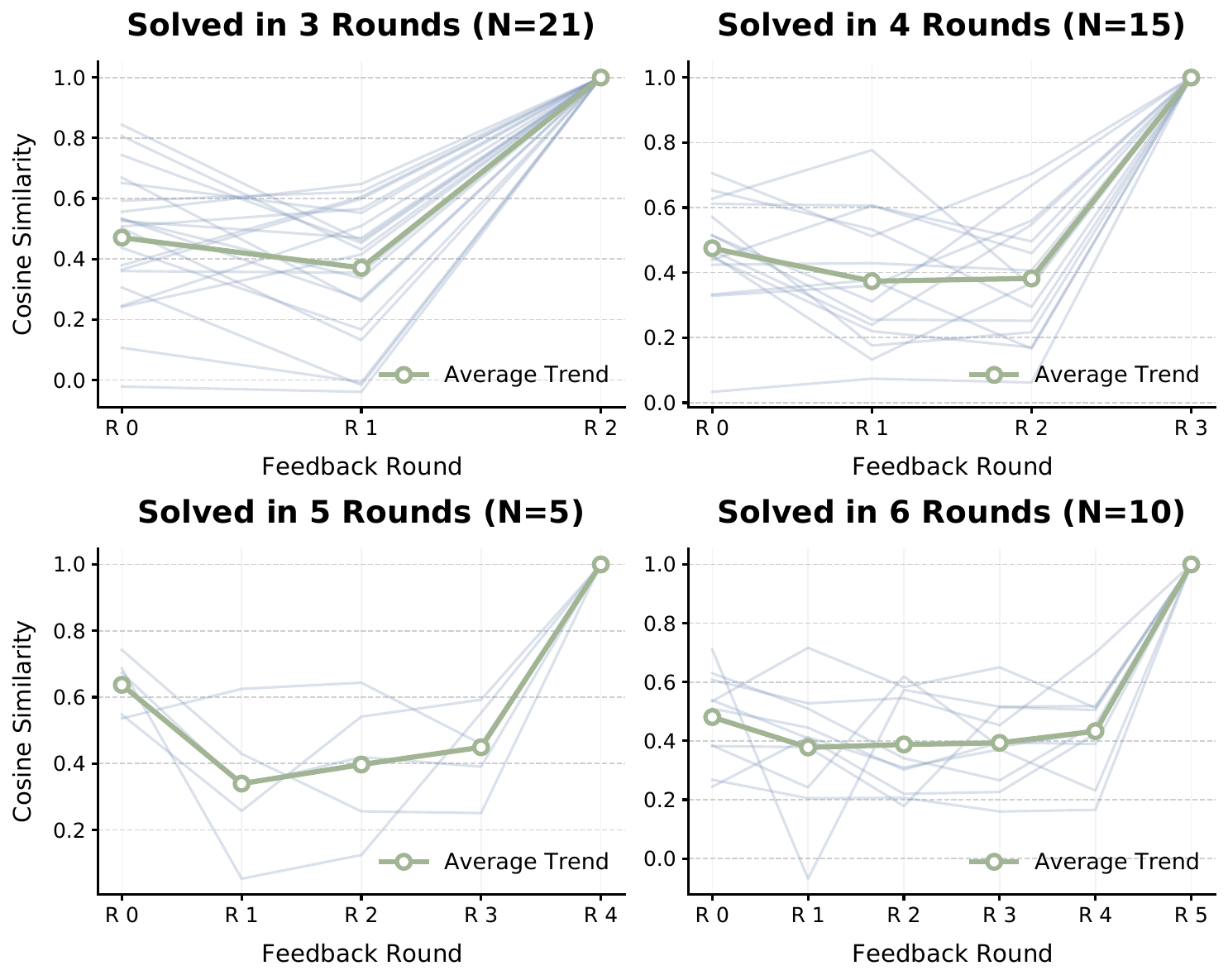}
\caption{Chinese MPAR-Bench feedback results for Qwen3-max in thinking mode, rounds 3--6. Blue lines show individual cases; green lines show average trends.}
\label{fig:word_emb_feedback}
\end{figure}

The proposed feedback method iteratively delivers semantic similarity metrics (ANLS and average word embedding similarity), guiding the model to refine its responses toward the answer across multiple rounds. Fig.~\ref{fig:word_emb_feedback} visualizes the corresponding trajectories (rounds 3–6) where the mechanism successfully corrected Qwen3-max's outputs. The results show that LLMs are not fully sensitive to word embedding similarity as a guidance signal; instead, they rely on other internal strategies. While feedback provides opportunities for exploration and sometimes enables recovery of the answer, the revision trajectory remains weakly aligned with semantic indicators, suggesting that models do not naturally exploit surface-level semantic proximity for iterative refinement.

\section{Conclusion}
We introduced \textbf{MPAR-Bench}, a bilingual benchmark that evaluates multi-point associative reasoning---reasoning \textit{breadth}---in LLMs through 1,000 boardgame-rule-based questions and a coarse-to-fine evaluation protocol spanning accuracy, ANLS, embedding similarity, and reasoning-trace verification, complemented by a four-axis perturbation suite.

Our experiments yield three findings. First, reasoning breadth remains far from solved: the best models reach 86.8\%/72.2\% accuracy in English/Chinese, with perturbations causing 9--18/5--12 point drops. Second, greater reasoning depth does not automatically confer breadth: thinking mode improves standard-setting accuracy but does not consistently reduce perturbation sensitivity, and case-level analysis reveals that extended reasoning can override correct answers through overthinking. Third, improving breadth appears challenging: scaling model size, adding reasoning strategies, and iterative feedback each bring only partial gains, suggesting that reasoning breadth may be a capability that current training paradigms do not naturally optimize for. We release MPAR-Bench and its pipeline to encourage the community to move beyond depth-oriented evaluation and toward a more complete picture of reasoning---one that values breadth as much as depth.

\newpage
\bibliography{references}

\appendix

\section*{Limitations}
While this benchmark provides an initial step toward evaluating the multi-point associative reasoning capabilities of LLMs, there remains some room to further broaden its coverage and ecological validity. Future work may extend the benchmark with more diverse and interactive settings to better capture associative reasoning behaviors that arise in real-world environments. In addition, we hope this benchmark can serve as a foundation for systematically studying non-linear associative and creative capabilities in LLMs, as well as for developing evaluation protocols and modeling principles that more closely align with practical applications of multi-point associative reasoning.

\section*{Ethical Considerations}
MPAR-Bench is built upon lexical materials sourced from cognitive psychology (the Remote Associates Test) and cooperative gaming (\textit{Just One}). Throughout dataset construction, we adopted a series of safeguards to preempt potential ethical risks. The benchmark consists solely of word-level clues and semantic associations, carrying no personally identifiable information and requiring no collection of human subject data. All clues were either drawn from publicly available game corpora or synthesized through an LLM-based multi-agent pipeline, after which embedding-based filtering and manual verification were applied to exclude content that could introduce demographic, cultural, or other unintended biases, in accordance with established ethical standards. The bilingual design of our benchmark further ensures equitable treatment of English and Chinese linguistic contexts, with dedicated attention to cultural appropriateness within each subset. Beyond its primary evaluation purpose, MPAR-Bench also serves as a diagnostic instrument for probing model reasoning behaviors and uncovering failure modes, thereby supporting the broader goal of identifying and mitigating ethical vulnerabilities in deployed AI systems.

\section{Multi-agent Generation Prompts}
\label{appendix:prompts}

Prompts are used in English and Chinese in generating, evaluating, and judging models separately.

\subsection{Questioner Prompt}
\label{appendix:questioner_prompt}
\begin{tcolorbox}[breakable,colback=gray!5,colframe=gray!75!black]
\small
You are playing the board game ``Just One''.

\vspace{1\baselineskip}

Secret answer: \{word\}

\vspace{1\baselineskip}

Requirements:

1. Give exactly 1 English clue word.

2. The clue must be a common English word or a widely recognized proper noun.

3. The clue must not be the answer itself, a translation, a synonym, a homophone, a made-up word, an obvious morphological variant, or contain the answer as a substring.

4. The clue should be indirect and moderately difficult, but useful when combined with other clues.

5. You MUST approach the answer from this specific association angle:  \{angle\}
\end{tcolorbox}

\subsection{Judger Prompt}
\label{appendix:judger_prompt}
\begin{tcolorbox}[breakable,colback=gray!5,colframe=gray!75!black]
\small
You are the judge for the board game ``Just One''.

\vspace{1\baselineskip}

Secret answer: \{word\}

Already approved clues (LOCKED - do NOT remove or re-evaluate these): \{locked\_clues\}

New candidate clues to evaluate: \{clue\_list\}.

Your tasks (apply only to the NEW candidates above):

1. Remove any new clue that IS the answer itself, a direct synonym, a translation, a homophone, an obvious morphological variant, or contains the full answer as a substring.

2. Remove EXACT duplicates among the new candidates.

3. Remove any new clue that refers to the SAME specific concept, entity, or phrase as an already-approved clue or another new candidate. Clues that merely belong to the same broad category (e.g., two different fictional characters, two different countries) are NOT duplicates - keep both.

4. Remove genuinely LOW-QUALITY new clues: completely obscure, made-up, grammatically wrong, or with no logical connection to the answer. Do NOT remove a clue just because it requires one step of reasoning.

5. Do NOT remove a clue simply because it seems ``too direct'' unless using it would immediately give away the answer with zero reasoning required.
\end{tcolorbox}

\subsection{Player Prompt}
\label{appendix:evaluate_prompt}
\begin{tcolorbox}[breakable,colback=gray!5,colframe=gray!75!black]
\small

You are the one who guess the mystery word for ``Just One''. 

\vspace{1\baselineskip}

Based on the clue words given, guess the mystery word and explain the logical connection between each related clue word and the mystery word.

The clue words are ``\{clue\_word1\}'', ``\{clue\_word2\}'', ``\{clue\_word3\}''

\vspace{1\baselineskip}

Requirements:

1. The mystery word is only one word.

2. Provide the connection between each clue word and the mystery word.
\end{tcolorbox}

\begin{table*}[!h]
\centering
\begin{tabular}{l|ccll}
\toprule
\textbf{Models} & \textbf{Model Size} & \textbf{Access} & \textbf{Version} & \textbf{Provider} \\
\midrule
GPT-5.2 & undisclosed & api & gpt-5.2-2025-12-11 & OpenAI \\
\midrule
Gemini-3.1pro & undisclosed & api & Gemini-3.1pro-preview & \multirow{2}{*}{Google} \\
Gemini-3flash & undisclosed & api & Gemini-3flash-preview & \\
\midrule
Sonnet-4.5 & undisclosed & api & claude-sonnet-4-5-20250929 & Anthropic \\
\midrule
Qwen3-max & undisclosed & api & qwen3-max-2026-01-23 & \multirow{7}{*}{Qwen} \\
Qwen3-0.6B & 0.6B & weights & - & \\
Qwen3-1.7B & 1.7B & weights & - & \\
Qwen3-4B & 4B & weights & - & \\
Qwen3-8B & 8B & weights & - & \\
Qwen3-14B & 14B & weights & - & \\
Qwen3-32B & 32B & weights & - & \\
\midrule
Kimi-k2 & 1T & api & kimi-k2-thinking-251104 & Moonshot AI \\
\midrule
Deepseek-v3.2 & 671B & api & Deepseek-v3.2 & DeepSeek \\
\midrule
Seed-2-pro & undisclosed & api & doubao-seed-2-0-pro-260215 & ByteDance \\
\bottomrule
\end{tabular}
\caption{Summary of Evaluated Models}
\label{tab:model_information}
\end{table*}

\begin{table}[!t]
\centering
\begin{tabular}{l l}
\toprule
\textbf{Component} & \textbf{Version} \\
\midrule
Python                  & 3.11.14 \\
PyTorch                 & 2.6.0 \\
transformers            & 4.57.1 \\
accelerate              & 1.10.1 \\
vllm                    & 0.16.0rc2 \\
openai                  & 1.96.1 \\
tiktoken                & 0.9.0  \\
fastText                & 0.9.3  \\
gensim                  & 4.3.0  \\
\bottomrule
\end{tabular}
\caption{Software versions used in our experiments}
\label{tab:software_versions}
\end{table}

\begin{table}[!h]
\centering
\begin{tabular}{lcc}
\toprule
Model & \makecell{Avg Token \\ (English)} & \makecell{Avg Token \\ (Chinese)} \\
\midrule
GPT-5.2 & 1355.1 & 1263.6 \\
Gemini-3.1pro & 3551.1 & 1883.3 \\
Sonnet-4.5 & 1468.9 & 1729.2 \\
Qwen3-max & 9983.3 & 777.9 \\
Kimi-k2 & 4659.5 & 3290.7 \\
Deepseek-v3.2 & 3091.0 & 2194.4 \\
Seed-2-pro & 1865.2 & 1563.8 \\
\bottomrule
\end{tabular}
\caption{Token Usage in Thinking Mode LLMs (Standard)}
\label{tab:token_usage_standard}
\end{table}
\begin{table}[!t]
\centering
\begin{tabular}{lcc}
\toprule
Model & \makecell{Avg Token \\ (English)} & \makecell{Avg Token \\ (Chinese)} \\
\midrule
GPT-5.2 & 1741.2 & 1338.9 \\
Gemini-3.1pro & 4940.7 & 2409.8 \\
Sonnet-4.5 & 1577.4 & 1838.7 \\
Qwen3-max & 11743.4 & 4782.4 \\
Kimi-k2 & 4790.1 & 4120.7 \\
Deepseek-v3.2 & 3255.0 & 2307.8 \\
Seed-2-pro & 2140.7 & 1858.6 \\
\bottomrule
\end{tabular}
\caption{Token Usage in Thinking Mode LLMs (Enhanced)}
\label{tab:token_usage_enhanced}
\end{table}

\section{Evaluation Prompt}
\label{appendix:cot_judge}
\begin{tcolorbox}[breakable,colback=gray!5,colframe=gray!75!black]
\small
You are a rigorous AI logic auditor. Your task is to evaluate each reasoning step produced by an Agent in a word-guessing game.

Break the reasoning into independent steps (atoms) and judge each atom on two dimensions: ``Factual Accuracy'' and ``Logical Soundness''.

\vspace{1\baselineskip}

For each step:

1. Fact\_Check: Are the objective claims in this step (numbers, ingredients, mechanisms, historical origins, etc.) factually correct? (Pass / Fail)

2. Logic\_Check: Is the reasoning chain from the clue to the predicted answer natural and sound? Are there signs of over-generalization, edge-case pandering, or multi-layer reinterpretation? (Pass / Fail)

\vspace{1\baselineskip}

Judging Principles (you MUST follow these):

- Scrutinize specific claims (numbers, ingredients, mechanisms, definitions) — any factual error → Fact Fail

- If a step cherry-picks a marginal meaning of the clue to fit the answer, ignoring a more obvious association → Logic Fail

- If the clue clearly points to a more specific/precise word, but the step generalizes it to a broad concept → Logic Fail

- If the step requires more than two layers of inference or reinterpretation to connect the clue to the answer → Logic Fail

- Only when the reasoning feels natural, direct, and requires no mental gymnastics for an ordinary person should it be judged Logic Pass
\end{tcolorbox}

\section{Asset Distribution and Compliance}
\label{app:assets}
To ensure reproducibility, both the English and Chinese subsets of MPAR-Bench, along with our evaluation scripts, will be publicly released under the MIT License upon publication. Our evaluation items are generated via frontier LLM APIs (e.g., GPT, Gemini, Qwen), and we have verified that our pipeline adheres to the respective terms of service of these model providers, restricting the usage of our dataset strictly to non-commercial academic benchmarking.

\begin{table}[!t]
\centering
\small
\setlength{\tabcolsep}{1.5pt}
\begin{tabular}{lcccc|cccc}
\toprule
\multirow{2}{*}{Model} &
\multicolumn{4}{c|}{English Acc(\%)} &
\multicolumn{4}{c}{Chinese Acc(\%)} \\
\cmidrule(r){2-5} \cmidrule(l){6-9}
& Mask & Shuf. & Dis. & \makecell{Multi\\step}
& Mask & Shuf. & Dis. & \makecell{Multi\\step} \\
\midrule
GPT-5.2 & 56.8 & 80.8 & 60.8 & 68.0 & 48.8 & 68.0 & \textbf{67.2} & 51.2 \\
Gemini-3.1pro & \textbf{67.2} & \textbf{86.3} & \textbf{76.7} & \textbf{77.6} & \textbf{55.2} & \textbf{73.6} & 66.4 & \textbf{60.8} \\
Sonnet-4.5 & 53.6 & 73.6 & 56.8 & 69.6 & 49.6 & 70.4 & \textbf{67.2} & 56.8 \\
Qwen3-max & 56.0 & 68.8 & 43.2 & 59.2 & 48.0 & 63.2 & 58.4 & 49.6 \\
Kimi-k2 & 49.6 & 69.6 & 49.6 & 61.6 & 35.2 & 57.6 & 48.8 & 41.6 \\
Deepseek-v3.2 & 40.8 & 64.8 & 44.8 & 57.6 & 44.8 & 60.0 & 57.6 & 47.2 \\
Seed-2-pro & 47.2 & 67.2 & 43.2 & 68.8 & 51.2 & 68.0 & 60.8 & 52.8 \\
\bottomrule
\end{tabular}
\caption{Detailed Results of Enhanced Benchmark on Thinking Models}
\label{tab:res_enhanced_think}
\end{table}

\begin{table}[t]
\centering
\small
\setlength{\tabcolsep}{1.5pt}
\begin{tabular}{lcccc|cccc}
\toprule
\multirow{2}{*}{Model} &
\multicolumn{4}{c|}{English Acc(\%)} &
\multicolumn{4}{c}{Chinese Acc(\%)} \\
\cmidrule(r){2-5} \cmidrule(l){6-9}
& Mask & Shuf. & Dis. & \makecell{Multi\\step}
& Mask & Shuf. & Dis. & \makecell{Multi\\step} \\
\midrule
GPT-5.2 & 32.0 & 54.5 & 39.2 & 51.2 & 42.4 & 67.2 & 64.8 & 43.2 \\
Gemini-3flash & 43.2 & 68.8 & \textbf{51.2} & \textbf{60.8} & 49.6 & \textbf{76.0} & 51.2 & \textbf{68.8} \\
Sonnet-4.5 & \textbf{45.6} & \textbf{72.8} & \textbf{51.2} & 53.6 & \textbf{53.6} & 71.2 & \textbf{71.2} & 52.0 \\
Qwen3-max & 35.2 & 55.2 & 44.0 & 44.8 & 44.0 & 65.6 & 67.2 & 47.2 \\
Deepseek-v3.2 & 32.0 & 56.8 & 39.2 & 42.4 & 47.2 & 61.6 & 62.4 & 44.8 \\
Seed-2-pro & 31.2 & 56.0 & 33.6 & 52.8 & 47.2 & 64.8 & 58.4 & 53.6 \\
\bottomrule
\end{tabular}
\caption{Detailed Results of Enhanced Benchmark on Non-thinking Models}
\label{tab:res_enhanced_non-thinking}
\end{table}

\section{Experiment Setup}
\label{appendix:experiment_setup}

For all evaluated models we keep the official default sampling parameters specified in each provider's release. For reasoning models, we additionally set reasoning\_effort=high where the API supports it. We do not perform any per-model hyper-parameter tuning. Detailed model configurations for all experiments are provided in
Table~\ref{tab:model_information}. And detailed software versions are provided in Table~\ref{tab:software_versions}.

\begin{table}[t]
\centering
\small
\setlength{\tabcolsep}{1pt}
\begin{tabular}{lcc|cc|cc|cc}
\toprule
\multirow{3}{*}{Model} &
\multicolumn{4}{c|}{English} &
\multicolumn{4}{c}{Chinese} \\
\cmidrule(r){2-5} \cmidrule(l){6-9}
&
\multicolumn{2}{c|}{Fact(\%)} &
\multicolumn{2}{c|}{Logic(\%)} &
\multicolumn{2}{c|}{Fact(\%)} &
\multicolumn{2}{c}{Logic(\%)} \\
\cmidrule(r){2-3} \cmidrule(r){4-5}
\cmidrule(r){6-7} \cmidrule(l){8-9}
& $\checkmark$ & $\times$ & $\checkmark$ & $\times$
& $\checkmark$ & $\times$ & $\checkmark$ & $\times$ \\
\midrule
GPT-5.2 & \textbf{1.75} & \textbf{6.61} & 6.39 & 24.64 & \textbf{0.87} & \textbf{2.25} & \textbf{5.59} & 11.69 \\
Gemini-3.1pro & 2.58 & 11.52 & \textbf{6.08} & \textbf{20.61} & 2.16 & 2.45 & 5.71 & \textbf{7.77} \\
Sonnet-4.5 & 4.30 & 17.14 & 8.71 & 31.62 & 3.44 & 5.28 & 8.55 & 14.72 \\
Qwen3-max & 4.37 & 18.96 & 8.03 & 38.21 & 2.22 & 4.94 & 9.22 & 15.30 \\
Kimi-k2 & 3.91 & 15.77 & 9.66 & 40.56 & 4.32 & 7.61 & 11.01 & 24.13 \\
Deepseek-v3.2 & 6.02 & 13.60 & 12.89 & 43.60 & 3.59 & 7.73 & 10.78 & 24.85 \\
Seed-2-pro & 5.27 & 20.14 & 7.90 & 34.69 & 2.48 & 5.08 & 7.06 & 13.33 \\
\bottomrule
\end{tabular}
\caption{Detailed Fact Fail and Logic Fail of Reasoning Trace on Standard MPAR-Bench on Thinking Models}
\label{tab:cot_standard_think}
\end{table}

\begin{table}[t]
\centering
\small
\setlength{\tabcolsep}{1pt}
\begin{tabular}{lcc|cc|cc|cc}
\toprule
\multirow{3}{*}{Model} &
\multicolumn{4}{c|}{English} &
\multicolumn{4}{c}{Chinese} \\
\cmidrule(r){2-5} \cmidrule(l){6-9}
&
\multicolumn{2}{c|}{Fact(\%)} &
\multicolumn{2}{c|}{Logic(\%)} &
\multicolumn{2}{c|}{Fact(\%)} &
\multicolumn{2}{c}{Logic(\%)} \\
\cmidrule(r){2-3} \cmidrule(r){4-5}
\cmidrule(r){6-7} \cmidrule(l){8-9}
& $\checkmark$ & $\times$ & $\checkmark$ & $\times$
& $\checkmark$ & $\times$ & $\checkmark$ & $\times$ \\
\midrule
GPT-5.2 & 5.30 & 24.06 & 9.93 & 47.03 & \textbf{1.75} & \textbf{3.87} & \textbf{6.34} & \textbf{12.88} \\
Gemini-3flash & 4.74 & 35.33 & \textbf{7.89} & 47.73 & 4.24 & 6.79 & 8.78 & 15.52 \\
Sonnet-4.5 & 5.40 & \textbf{9.94} & 27.30 & \textbf{45.41} & 3.84 & 7.18 & 9.24 & 14.10 \\
Qwen3-max & \textbf{4.55} & 31.30 & 8.52 & 45.65 & 2.61 & 4.49 & 8.32 & 15.28 \\
Deepseek-v3.2 & 8.95 & 29.71 & 15.10 & 55.97 & 2.18 & 8.83 & 8.84 & 21.83 \\
Seed-2-pro & 10.03 & 37.51 & 15.65 & 52.14 & 3.43 & 8.27 & 8.47 & 20.34 \\
\bottomrule
\end{tabular}
\caption{Detailed Fact Fail and Logic Fail of Reasoning Trace on Standard MPAR-Bench on Non-thinking Models}
\label{tab:cot_standard_non-thinking}
\end{table}

\begin{table}[t]
\centering
\small
\setlength{\tabcolsep}{1pt}
\begin{tabular}{lcc|cc|cc|cc}
\toprule
\multirow{3}{*}{Model} &
\multicolumn{4}{c|}{English} &
\multicolumn{4}{c}{Chinese} \\
\cmidrule(r){2-5} \cmidrule(l){6-9}
&
\multicolumn{2}{c|}{Fact(\%)} &
\multicolumn{2}{c|}{Logic(\%)} &
\multicolumn{2}{c|}{Fact(\%)} &
\multicolumn{2}{c}{Logic(\%)} \\
\cmidrule(r){2-3} \cmidrule(r){4-5}
\cmidrule(r){6-7} \cmidrule(l){8-9}
& $\checkmark$ & $\times$ & $\checkmark$ & $\times$
& $\checkmark$ & $\times$ & $\checkmark$ & $\times$ \\
\midrule
GPT-5.2 & \textbf{2.52} & \textbf{6.22} & 10.38 & 26.85 & \textbf{2.27} & \textbf{2.30} & \textbf{9.94} & 13.31 \\
Gemini-3.1pro & 4.07 & 10.71 & \textbf{9.85} & \textbf{23.51} & 3.11 & 3.59 & 10.97 & \textbf{12.56} \\
Sonnet-4.5 & 6.79 & 16.49 & 12.93 & 39.21 & 6.11 & 8.35 & 15.73 & 22.32 \\
Qwen3-max & 4.74 & 15.78 & 11.86 & 35.90 & 3.15 & 4.30 & 11.83 & 17.13 \\
Kimi-k2 & 5.96 & 14.03 & 14.02 & 40.25 & 5.61 & 10.15 & 18.18 & 25.07 \\
Deepseek-v3.2 & 7.09 & 14.37 & 17.11 & 45.32 & 5.97 & 9.80 & 17.11 & 31.04 \\
Seed-2-pro & 7.07 & 21.60 & 12.21 & 37.82 & 5.83 & 7.00 & 14.87 & 18.23 \\
\bottomrule
\end{tabular}
\caption{Detailed Fact Fail and Logic Fail of Reasoning Trace on Enhanced MPAR-Bench on Thinking Models}
\label{tab:cot_enhanced_think}
\end{table}

\begin{table}[t]
\centering
\small
\setlength{\tabcolsep}{1pt}
\begin{tabular}{lcc|cc|cc|cc}
\toprule
\multirow{3}{*}{Model} &
\multicolumn{4}{c|}{English} &
\multicolumn{4}{c}{Chinese} \\
\cmidrule(r){2-5} \cmidrule(l){6-9}
&
\multicolumn{2}{c|}{Fact(\%)} &
\multicolumn{2}{c|}{Logic(\%)} &
\multicolumn{2}{c|}{Fact(\%)} &
\multicolumn{2}{c}{Logic(\%)} \\
\cmidrule(r){2-3} \cmidrule(r){4-5}
\cmidrule(r){6-7} \cmidrule(l){8-9}
& $\checkmark$ & $\times$ & $\checkmark$ & $\times$
& $\checkmark$ & $\times$ & $\checkmark$ & $\times$ \\
\midrule
GPT-5.2 & 7.16 & \textbf{24.19} & 13.97 & 47.39 & 3.38 & \textbf{5.73} & \textbf{10.11} & \textbf{19.29} \\
Gemini-3flash & 7.22 & 33.00 & 13.62 & 48.06 & 6.38 & 11.22 & 13.60 & 24.33 \\
Sonnet-4.5 & 8.52 & 26.70 & 14.79 & \textbf{46.04} & 5.97 & 10.44 & 15.73 & 25.45 \\
Qwen3-max & \textbf{6.25} & 25.56 & \textbf{13.21} & 48.91 & \textbf{3.32} & 9.29 & 14.47 & 25.21 \\
Deepseek-v3.2 & 11.79 & 29.63 & 20.09 & 56.30 & 4.58 & 11.50 & 14.49 & 30.75 \\
Seed-2-pro & 12.97 & 33.95 & 18.03 & 50.88 & 5.36 & 11.36 & 14.08 & 24.81 \\
\bottomrule
\end{tabular}
\caption{Detailed Fact Fail and Logic Fail of Reasoning Trace on Enhanced MPAR-Bench on Non-thinking Models}
\label{tab:cot_enhanced_non-thinking}
\end{table}

\section{Detailed Results}
\label{appendix:detailed_results}

\subsection{Token Usage}

Tables~\ref{tab:token_usage_standard} and~\ref{tab:token_usage_enhanced} report the average token consumption of models operating in thinking mode, encompassing prompt tokens, Chain-of-Thought reasoning tokens, and output tokens.

Token consumption increases consistently from the Standard to the Enhanced setting for nearly all models, reflecting the greater reasoning demand imposed by the four perturbation conditions. The magnitude of this increase, however, varies markedly across models, pointing to fundamentally different strategies for allocating reasoning computation.

The most pronounced outlier is Qwen3-max, which exhibits disproportionately high token consumption in English across both settings, averaging roughly 10,000 tokens in Standard MPAR-Bench and 11,700 in Enhanced MPAR-Bench. Qualitative inspection of its generated reasoning trace reveals a consistent tendency toward repetitive self-verification: the model frequently revisits intermediate conclusions, generates multiple redundant candidate answers, and enters circular deliberation loops before converging on a final prediction. This overthinking behavior substantially inflates token usage.

\subsection{Enhanced MPAR-Bench Result Analysis}
\label{appendix:enhanced_mpar_bench_analysis}
Tables~\ref{tab:res_enhanced_think} and~\ref{tab:res_enhanced_non-thinking} present the detailed accuracy of model across four perturbation variants in the Enhanced MPAR-Bench setting: Clue Masking, Order Shuffling, Distractor Injection, and Multi-step Inferring.

The four perturbation types exhibit different impacts on model performance. Order Shuffling consistently yields the highest accuracy across all models, some of which are even higher than Standard setting. In contrast, Clue Masking causes the most severe degradation in English, with an average drop of 20.0\% from Order Shuffling. Distractor Injection also substantially reduces performance, particularly for Qwen3-max and Seed-2-pro, both dropping by over 28\% from their respective Standard accuracies. This indicates that spurious semantic correlations introduced by irrelevant clue words can effectively derail the model's reasoning trajectory. Multi-step Inferring occupies an intermediate difficulty level, suggesting that extending the associative chain length moderately taxes the model's long-range semantic mapping capacity but does not fundamentally break the reasoning process.

\subsection{Reasoning Trace Error Analysis}
\label{appendix:cot_error_analysis}

Tables~\ref{tab:cot_standard_think} through~\ref{tab:cot_enhanced_non-thinking} present the reasoning trace evaluation results, decomposing reasoning failures into ``fact fail'' and ``logic fail''.

Across all models, logical error rates substantially exceed factual error rates. In the English thinking standard setting, Deepseek-v3.2 exhibits a logical error rate of 43.60\%, Kimi-k2 40.56\%, and Qwen3-max 38.21\%. Even the best-performing model, Gemini-3.1pro, shows a logical error rate of 20.61\%. By contrast, factual error rates in the same setting are considerably lower: Deepseek-v3.2 at 13.60\%, Kimi-k2 at 15.77\%, and Gemini-3.1pro at 11.52\%. This indicates that the primary failure in multi-point associative reasoning is not incorrect factual knowledge but rather invalid inferential jumps in constructing the reasoning chain from clues to the answer. Moreover, logical error rates in Chinese reasoning are consistently lower than in English.

Furthermore, in the non-thinking mode, factual error rates increase substantially, which suggests that without an explicit reasoning process, models are more susceptible to factual hallucination.

\subsection{Structured Reasoning Skill Experiment}
\label{appendix:skill_experiment}

We designed a structured three-step reasoning skill to test whether explicit multi-step prompting can enhance reasoning breadth beyond what thinking mode alone achieves. The skill instructs the model to: (1) examine all clues comprehensively and identify their underlying associations; (2) prioritize specific concepts over abstract hypernyms; and (3) perform reverse verification by reasoning backward from the predicted answer to the provided clues. We evaluate this skill on Seed-2-pro (thinking mode, standard setting).

On the English subset, accuracy rises marginally from 71.4\% to 72.4\% (+1.0pp), ANLS from 0.724 to 0.734, and embedding similarity from 0.802 to 0.807. On the Chinese subset, gains are somewhat larger: accuracy from 64.6\% to 67.8\% (+3.2pp), ANLS from 0.739 to 0.765, and embedding similarity from 0.820 to 0.838. The modest magnitude of improvement---particularly on English---suggests that thinking-mode models may already perform these associative integration steps implicitly during extended reasoning, and that prompt-level interventions alone offer limited leverage on the core challenge of multi-source evidence integration.

\end{document}